\documentclass[11pt,a4paper]{article}
\usepackage[hyperref]{acl2020}
\usepackage{times}
\usepackage{latexsym}
\usepackage{CJKutf8}
\usepackage{graphicx}
\usepackage{amsmath}
\usepackage{amssymb}
\usepackage{multirow}
\usepackage{booktabs}

\usepackage{microtype}

\aclfinalcopy 

\title{Unified Multi-Criteria Chinese Word Segmentation with BERT}

\author{
  Zhen Ke\textsuperscript{1,2}, Liang Shi\textsuperscript{2}, Erli Meng\textsuperscript{2}, Bin Wang\textsuperscript{2}, Xipeng Qiu\textsuperscript{1}, Xuanjing Huang\textsuperscript{1} \\
  Shanghai Key Laboratory of Intelligent Information Processing, Fudan University\textsuperscript{1}\\
  Xiaomi AI Lab, Xiaomi Inc., Beijing, China\textsuperscript{2}\\
  \texttt{\{zke17,xpqiu,xjhuang\}@fudan.edu.cn} \\
  \texttt{\{shiliang1,mengerli,wangbin11\}@xiaomi.com} \\
}

\date{}

\begin{document}
\begin{CJK}{UTF8}{gbsn}
\maketitle
\begin{abstract}
Multi-Criteria Chinese Word Segmentation (MCCWS) aims at finding word boundaries in a Chinese sentence composed of continuous characters while multiple segmentation criteria exist.
The unified framework has been widely used in MCCWS and shows its effectiveness.
Besides, the pre-trained BERT language model has been also introduced into the MCCWS task in a multi-task learning framework.
In this paper, we combine the superiority of the unified framework and pre-trained language model, and propose a unified MCCWS model based on BERT.
Moreover, we augment the unified BERT-based MCCWS model with the bigram features and an auxiliary criterion classification task.
Experiments on eight datasets with diverse criteria demonstrate that our methods could achieve new state-of-the-art results for MCCWS.
\end{abstract}

\section{Introduction}
\label{sec:intro}

Chinese Word Segmentation (CWS) is a fundamental task in Chinese Natural Language Processing (NLP), which aims to identify word boundaries in a Chinese sentence consisting of continuous characters.
Most approaches transform the CWS problem into a character-based sequence labeling problem, in which each character is assigned a label to denote its position in the target word \citep{xue-2003-chinese, zheng-2013-deep, chen-2015-gated, chen-2015-long, zhang-2016-transition, ma-2018-state, yang-2019-subword}.
Recently, the CWS task of multiple segmentation criteria is studied, which can be formulated as Multi-Criteria Chinese Word Segmentation (MCCWS) task \citep{chen-2017-adversarial, he-2019-effective, gong-2019-switch, huang-2019-toward, qiu-2019-multi}.
In MCCWS, a sentence can be segmented into different word units in different criteria.
An example of MCCWS is showed in Table \ref{tbl:example}.

\begin{table}[htbp] \small
\centering
\begin{tabular}{|c|c|c|c|c|c|}
\hline
Criteria & Li & Na & entered & \multicolumn{2}{c|}{the semi-final} \\ \hline
CTB & \multicolumn{2}{c|}{李娜} & 进入 & \multicolumn{2}{c|}{半决赛} \\ \hline
PKU & 李 & 娜 & 进入 & 半 & 决赛 \\ \hline
MSRA & \multicolumn{2}{c|}{李娜} & 进入 & 半 & 决赛 \\ \hline
\end{tabular}
\caption{An example of MCCWS.}
\label{tbl:example}
\end{table}

Existing MCCWS methods can be divided into two frameworks: the multi-task learning framework and the unified framework.
The multi-task learning framework treats MCCWS as a multi-task learning problem \citep{caruana-1997-multitask}, with each criterion as a single task \citep{chen-2017-adversarial, huang-2019-toward}.
A shared layer is used to obtain the common features of all criteria, and a private layer of each criterion obtains the criterion-specific features.
The multi-task learning framework can learn shared knowledge but not efficiently, which suffers from high model complexity and complex training algorithm.
The unified framework employs a fully shared model for all criteria in MCCWS, and the criterion is fed as input \citep{he-2019-effective, gong-2019-switch, qiu-2019-multi}.
The unified framework is more concise, in which the shared knowledge are fully used and training process is simplified, compared with the multi-task learning framework.

Recently, the pre-trained BERT language model \citep{devlin-2019-bert} have showed its powerful capability to learn prior linguistic knowledge, which have been proved beneficial for many NLP tasks. \citet{huang-2019-toward} also introduced BERT to MCCWS in the multi-task learning framework and achieved excellent performance. However, due to the powerful ability of BERT, it is unnecessary to use multiple criterion-specific projection layers, which motivates us to explore BERT for MCCWS in the more concise unified framework.

In this paper, we propose a BERT-based unified MCCWS model, to incorporate the superiority of unified framework and pre-trained language model for the first time.
Besides, the bigram features have been proved to be effective for CWS~\cite{chen-2015-long, zhang-2016-transition}. Thus, a fusion layer and an attention layer are further employed to integrate the bigram features and the output of BERT.
To ensure the unified model retrains the criterion information, an auxiliary criterion classification task is introduced to further boosts the performance. Different to the previous work, our model does not use the CRF~\cite{lafferty-2001-conditional} inference layer, thereby being more efficient in inference phase by fully utilizing the ability of parallel computation.
Experiments show that our model can achieve new state-of-the-art results on eight standard datasets.

\section{Methodology}
\label{sec:method}

As previous work \citep{chen-2017-adversarial, huang-2019-toward, qiu-2019-multi}, the MCCWS task is viewed as a character-based sequence labeling problem.
Specifically, given a Chinese sentence $ X = \{ x_1, x_2, ..., x_T \} $ composed of continuous characters and a criteria $c$, the model should output a CWS label sequence $ Y = \{ y_1, y_2, ..., y_T \} $ with $ y_t \in \{ B, M, E, S \} $, denoting the beginning, middle, end of a word, or a word as single character.

\begin{figure}[htbp]
    \centering
    \includegraphics[width=0.43\textwidth]{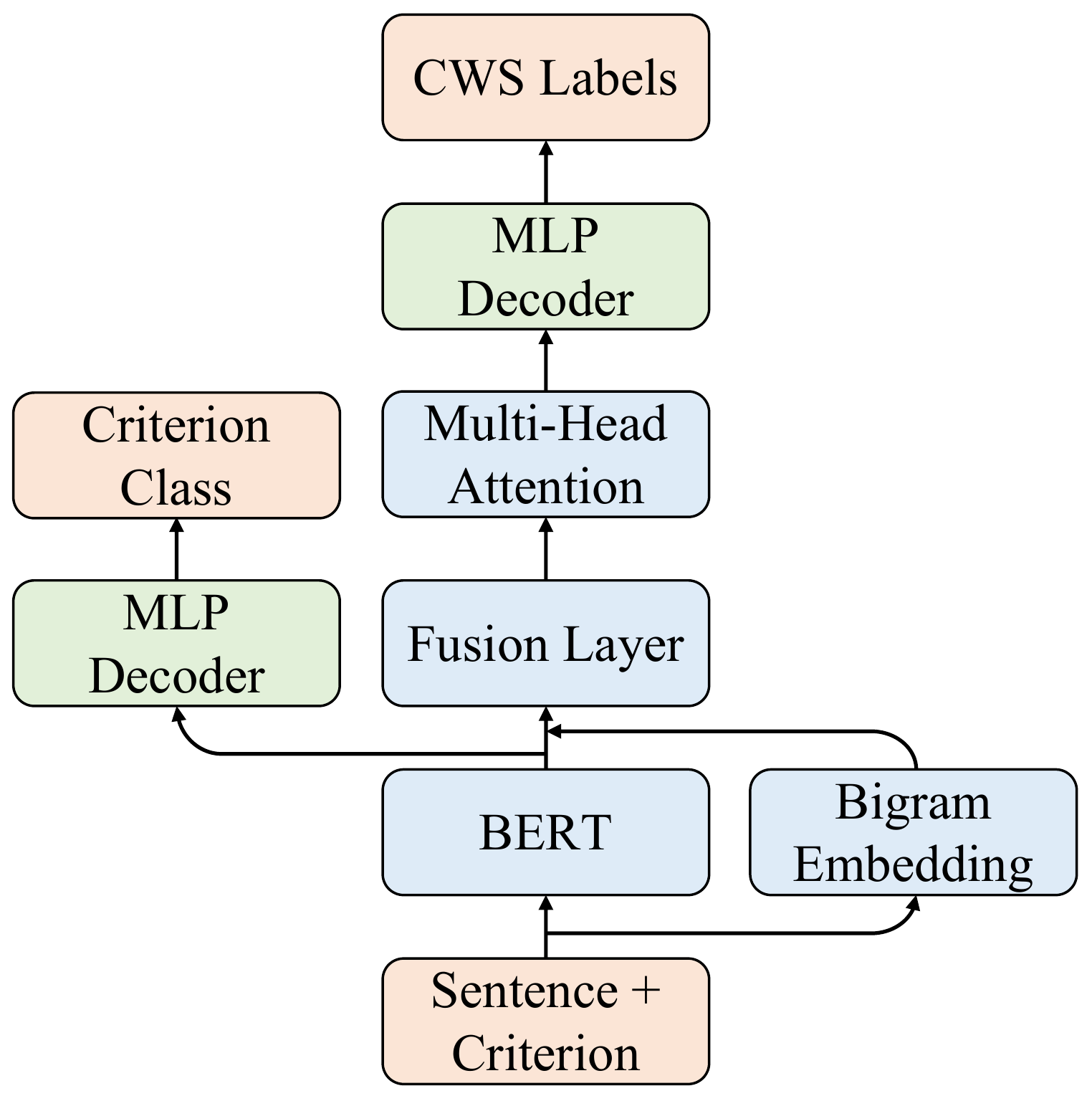}
    \caption{Overall architecture of our proposed model.}
    \label{fig:model}
\end{figure}

In our model, we first augment the input sentence by inserting a specific criterion token before it, and encode the augmented sentence using the BERT model \citep{devlin-2019-bert}.
Besides, we convert character-based bigram features into bigram embedding.
Then, we use a fusion gate mechanism to integrate the hidden representation from BERT with the bigram embedding.
A multi-head attention is followed to interchange context information among the fused representations.
Finally, a multi-layer Perceptron (MLP) decoder is used to decode the segmentation labels.
Besides, an auxiliary criterion classifier is employed to reconstruct the criterion class from the input.
The overall architecture of our model is displayed in Figure \ref{fig:model}.

\subsection{Augmented Input Sentence}
\label{ssec:augment}

We add a specific criterion token before the original input sentence $X$ to obtain the augmented sentence $X' = \{ x_0, x_1, .., x_T \}$.
For example, we add the token \texttt{<pku>} before the original sentence indicating that it adheres to the PKU criterion.

\subsection{Encoder}
\label{ssec:encoder}

\paragraph{BERT}
BERT \citep{devlin-2019-bert} is short for Bidirectional Encoder Representations from Transformers, which is a Transformer \citep{vaswani-2017-attention} based bidirectional language model pre-trained on a large-scale unsupervised corpus.
We employ BERT as our basic encoder for MCCWS to introduce prior linguistic knowledge, converting the augmented sentence to hidden character representations:
\begin{equation} \small
    \mathbf{H} = \text{BERT} (X') ,
    \label{eq:bert}
\end{equation}
where $\mathbf{H} \in \mathbb{R}^{(T+1) \times d_h}$.

\paragraph{Bigram Embedding}
The bigram features have proved beneficial for MCCWS \citep{chen-2017-adversarial, he-2019-effective, qiu-2019-multi}.
Therefore, we construct the bigram feature for every character by concatenating it with the previous character, $ B = \{ x_0 x_1, x_1 x_2, ..., x_{T-1} x_T \} $.
Then, we convert the bigram features to bigram embedding vectors by looking them up in an embedding table
\begin{equation} \small
    \mathbf{E} = \text{BigramEmbed} (B),
    \label{eq:bi-embed}
\end{equation}
where $\mathbf{E} \in \mathbb{R}^{T \times d_e}$.

\paragraph{Fusion Layer}
We use a fusion gate mechanism to integrate the hidden character representations and the bigram embedding representations.
We refer to the $t$-th hidden character vector as $\mathbf{h}_t \in \mathbb{R}^{d_h}$, the $t$-th bigram embedding as $\mathbf{e}_t \in \mathbb{R}^{d_e}$.
The fusion gate mechanism can be formulated as follows,
\begin{equation} \small
\begin{aligned}
    \mathbf{h}'_t &= \tanh \left( \mathbf{W}_h \mathbf{h}_t + \mathbf{b}_h \right), \\
    \mathbf{e}'_t &= \tanh \left( \mathbf{W}_e \mathbf{e}_t + \mathbf{b}_e \right), \\
    \mathbf{g}_t &= \sigma \left( \mathbf{W}_{fh} \mathbf{h}_t + \mathbf{W}_{fe} \mathbf{e}_t + \mathbf{b}_f \right), \\
    \mathbf{f}_t &= \mathbf{g}_t \odot \mathbf{h}'_t + \left( 1 - \mathbf{g}_t \right) \odot \mathbf{e}'_t,
\end{aligned}
\label{eq:fusion}
\end{equation}
where $ \mathbf{h}'_t, \mathbf{e}'_t, \mathbf{g}_t, \mathbf{f}_t \in \mathbb{R}^{d_h}$, $\mathbf{g}_t$ is the fusion gate vector, $\mathbf{f}_t$ is the fused vector, and $\sigma$ is the sigmoid function.

\paragraph{Multi-Head Attention}
Next, we employ a multi-head attention \citep{vaswani-2017-attention} to obtain the contextual representations.
It is necessary to contextualize the fused representations because the fusion layer can only integrate the unigram and bigram representations character-wise, lacking the knowledge in the context.
The multi-head attention with residual connection \citep{he-2016-deep} and layer normalization \citep{ba-2016-layer} can be formulated as
\begin{equation} \small
    \mathbf{O} = \text{LayerNorm} \left( \text{MultiHead} (\mathbf{F}) + \mathbf{F} \right),
\label{eq:multi-head}
\end{equation}
where $\mathbf{F} \in \mathbb{R}^{T \times d_h}$ is the fused representations and $\mathbf{O} \in \mathbb{R}^{T \times d_h}$ is the final output representations from the encoder.

\subsection{Decoder}
\label{ssec:decoder}

\paragraph{CWS Label Decoder}
The output representations are converted into the probabilities over the CWS labels by an MLP layer,
\begin{equation} \small
\begin{aligned}
    P(y_t | X, c) &= \text{softmax} \left( \mathbf{W}_o \mathbf{o}_t + \mathbf{b}_o \right)_{y_t} \\
    P(Y|X,c) &= \sum_{t=1}^T P(y_t | X, c)
\end{aligned}
\label{eq:label}
\end{equation}
where $\mathbf{W}_o \in \mathbb{R}^{4 \times d_h}$.
$P(y_t | X, c)$ means the probability that the $t$-th CWS label is $y_t$, and $P(Y|X,c)$ means the probability of the label sequence $Y$, given the input sentence $X$ and the criterion $c$.

\paragraph{Auxiliary Criterion Classifier}
To avoid the criterion information lost in forward propagation, we add an auxiliary criterion classifier to reconstruct the criterion from the hidden representations.
The probability of reconstructing criterion $c$ is
\begin{equation}\small
    P(c|X,c) = \text{softmax} \left( \mathbf{W}_c \mathbf{h}_0 + \mathbf{b}_c \right)_c ,
\label{eq:criteria}
\end{equation}
where $\mathbf{W} \in \mathbb{R}^{C \times d_h}$ and $C$ is the number of criteria.

\subsection{Loss}
\label{ssec:loss}

We use the negative log likelihood objective as our loss function, and add the losses for CWS labeling and criterion classification directly
\begin{equation} \small
    L = - \sum_{i=1}^N \log P(Y_i|X_i,c_i) + \log P(c_i|X_i,c_i) ,
\end{equation}
where $N$ is the number of training samples.

\section{Experiments}
\label{sec:experiment}

\subsection{Datasets}
\label{ssec:dataset}

We experiment on eight CWS datasets from SIGHAN2005 \citep{emerson-2005-second} and SIGHAN2008 \citep{jin-2008-fourth}, among which MSRA, PKU, CTB, NCC and SXU are simplified Chinese datasets, while AS, CKIP and CITYU are traditional Chinese datasets.
As previous work \citep{he-2019-effective, qiu-2019-multi}, we convert all traditional Chinese datasets into simplified Chinese using the OpenCC library \footnote{\url{https://github.com/BYVoid/OpenCC}}.
All datasets are preprocessed by converting all full-width digits, punctuation and Latin letters to half-width, and replacing the continuous English characters and digits with a specific token respectively.


\subsection{Settings}
\label{ssec:setting}

We use the Chinese BERT model pre-trained with whole word masking (BERT-wwm) \citep{cui-2019-pre-training}, whose number of layers is 12, number of heads if 12, hidden size is $d_h=768$.
We use the bottom 6 layers as our BERT model for the balance of performance and speed.
The size of bigram embeddings is set to be $d_e=100$ and the bigram embeddings are pre-trained on the Chinese Wikipedia corpus as \citet{qiu-2019-multi}.
The dropout probability of all hidden layers is set to be 0.1.

The AdamW \citep{loshchilov-2019-decoupled} optimizer is used in our fine-tuning process, with $\beta_1 = 0.9$, $\beta_2 = 0.999$ and weight decay of 0.01.
The initial learning rate is 2e-5, and a linear warm-up of ratio 0.1 is used to adjust the learning rate dynamically.
We set the number of epochs to 10 and the batch size to 64. Our model is implemented using the fastNLP library \footnote{\url{https://github.com/fastnlp/fastNLP}}.

\subsection{Overall Results}
\label{ssec:result}

Table \ref{tbl:result} shows results of on the test sets of eight CWS dataset, with F1 values as metrics.
The results are displayed in four blocks.

\begin{table*}[htbp] \small
\centering
\begin{tabular}{lccccccccc}
\toprule
Method & MSRA & AS & PKU & CTB & CKIP & CITYU & NCC & SXU & Avg. \\ \toprule
Bi-LSTM & 95.84 & 94.20 & 93.30 & 95.30 & 93.06 & 94.07 & 92.17 & 95.17 & 94.14 \\ \midrule
Adversarial & 96.04 & 94.64 & 94.32 & 96.18 & 94.26 & 95.55 & 92.83 & 96.04 & 94.98 \\
Multi-Task BERT & 97.9 & 96.6 & 96.6 & - & - & \textbf{97.6} & - & 97.3 & - \\ \midrule
Unified Bi-LSTM & 97.35 & 95.47 & 95.78 & 95.84 & 95.73 & 95.60 & 94.34 & 96.49 & 95.73 \\
Switch-LSTMs & 97.78 & 95.22 & 96.15 & 97.26 & 94.99 & 96.22 & 94.12 & 97.25 & 96.12 \\
Transformer & 98.05 & 96.44 & 96.41 & 96.99 & 96.51 & 96.91 & 96.04 & 97.61 & 96.87 \\ \midrule
Unified BERT & 98.45 & \textbf{96.90} & 96.89 & \textbf{97.20} & 96.83 & 97.07 & 96.48 & \textbf{97.81} & \textbf{97.204} \\
\ \ - Bigram & 98.38 & 96.88 & 96.87 & 97.14 & 96.72 & 97.05 & 96.33 & 97.74 & 97.139 \\
\ \ - CLS & \textbf{98.48} & 96.86 & \textbf{96.92} & 97.13 & \textbf{96.84} & 97.07 & \textbf{96.55} & 97.72 & 97.196 \\
\ \ - CLS - Bigram  & 98.41 & 96.83 & 96.83 & 97.13 & 96.76 & 97.05 & 96.33 & 97.67 & 97.126 \\
\bottomrule
\end{tabular}
\caption{
F1 values on test sets of eight standard CWS datasets.
There are four blocks, indicating single criterion methods, multi-task learning framework methods, unified framework methods, and our methods, respectively.
The maximum F1 values are highlighted for each dataset.
}
\label{tbl:result}
\end{table*}

\begin{table*}[ht] \small
\centering
\begin{tabular}{lccccccccc}
\toprule
Method & MSRA & AS & PKU & CTB & CKIP & CITYU & NCC & SXU & Avg. \\ \midrule
Bi-LSTM & 66.28 & 70.07 & 66.09 & 76.47 & 72.12 & 65.79 & 59.11 & 71.27 & 68.40 \\
Adversarial & 71.60 & 73.50 & 72.67 & 82.48 & 77.59 & 81.40 & 63.31 & 77.10 & 74.96 \\
Switch-LSTMs & 64.20 & 77.33 & 69.88 & 83.89 & 77.69 & 73.58 & 69.76 & 78.69 & 74.38 \\
Transformer & 78.92 & 76.39 & 78.91 & 87.00 & 82.89 & 86.91 & 79.30 & 85.08 & 81.92 \\
Multi-Task BERT & \textbf{84.0} & 76.9 & \textbf{80.1} & - & - & \textbf{89.7} & - & 86.0 & - \\
Unified BERT& 83.35 & \textbf{79.26} & 79.71 & \textbf{87.77} & \textbf{84.36} & 87.27 & \textbf{81.03} & \textbf{86.05} & \textbf{83.60} \\ \bottomrule
\end{tabular}
\caption{OOV recalls on test sets of eight CWS datasets. The best results are highlighted for each dataset.}
\label{tbl:oov}
\end{table*}

The first block shows the single criterion methods for CWS.
\textbf{Bi-LSTM} \citep{chen-2017-adversarial} is the baseline method trained on each dataset separately.

The second block shows the multi-task learning framework methods.
\textbf{Adversarial} \citep{chen-2017-adversarial} constructs a multi-task model which adopts an adversarial strategy to learn more criteria-invariant representations.
\textbf{Multi-task BERT} \citep{huang-2019-toward} uses pre-trained BERT as the sentence encoder.

The third block shows the unified framework methods.
\textbf{Unified Bi-LSTM} \citep{he-2016-deep} introduces two artificial tokens at the beginning and ending of input sentence to specify the target criterion.
\textbf{Switch-LSTMs} \citep{gong-2019-switch} uses Switch-LSTMs as the backbone network.
\textbf{Transformer} \citep{qiu-2019-multi} uses the Transformer network \citep{vaswani-2017-attention} as the basic encoder.

The fourth block shows our methods for MCCWS.
\textbf{Unified BERT} is the complete model discussed in Section \ref{sec:method}, which incorporates the unified framework and the pre-trained BERT model. It is also augmented by the fused bigram features and the auxiliary criterion classification task.
Besides, \textbf{- Bigram} is the \textbf{Unified BERT} model without bigram features, while the \textbf{- CLS} model is the model without criterion classification and \textbf{- CLS - Bigram} is the model without bigram features and criterion classification.

From Table \ref{tbl:result}, we could see that our proposed methods outperform previous methods obviously on nearly all datasets, especially two previous best methods: BERT based multi-task method \textbf{Multi-task BERT} and unified method \textbf{Transformer}.
These promotions should be attributed to the prior knowledge introduced by BERT and the shared knowledge captured by the unified framework.
Besides, we could see that the complete \textbf{Unified BERT} model outperforms the model without bigram features (\textbf{- Bigram}) and the model without auxiliary criterion classification (\textbf{- CLS}), which shows the benefit of bigram features and criterion classification.
The complete model obtains the best average performance and achieve new state-of-the-art results for MCCWS, showing the effectiveness of our method.

\subsection{OOV Words}
\label{ssec:oov}

OOV words means Out-of-Vocabulary words.
According to previous work \citep{ma-2018-state, huang-2019-toward}, OOV error is a critical contribution to the total error of CWS.
We use OOV recall as metrics to evaluate the performance on OOV words.
The results are showed in Table \ref{tbl:oov}.

We can see that our proposed method can achieve best OOV performance on five datasets, and comparable performance on the other three datasets.
The results demonstrate the effectiveness of our method on OOV words.

\section{Conclusion}
\label{sec:conclusion}

In this paper, we mainly focus on the Multi-Criteria Chinese Word Segmentation (MCCWS) task. We propose a new method which incorporates the superiority of the unified framework and the pre-trained BERT model. Augmented with bigram features and an auxiliary criterion classification task, we achieve the new state-of-the-art results for MCCWS.


\bibliography{acl2020}
\bibliographystyle{acl_natbib}

\end{CJK}
\end{document}